\let\oldvec\vec
\documentclass[oribibl,graybox]{svmult}
\let\vec\oldvec
\usepackage{cite}
\usepackage{graphicx}
\usepackage{subfigure}
\usepackage{algorithm2e}
\usepackage{amsmath}
\usepackage[numbers]{natbib}
\usepackage{balance}
\usepackage{multirow}
\usepackage{float}
\usepackage{stfloats}
\usepackage{flushend}
\usepackage{amssymb}
\usepackage{float}
\usepackage{mathptmx}       
\usepackage{helvet}         
\usepackage{courier}        
\usepackage{type1cm}        
\newcommand\textlcsc[1]{\textsc{\MakeLowercase{#1}}}             
\usepackage{makeidx}         
\usepackage{graphicx}        
\usepackage{multicol}        
\usepackage[bottom]{footmisc}

\makeindex             

\begin{document}
\title*{\textlcsc{Multi-level SVM Based CAD Tool for Classifying Structural MRI}s}
\author{Jerrin Thomas Panachakel  \and Jeena R.S.}
\institute{Jerrin Thomas Panachakel\at
	Dept. of Electrical Engineering \\
	Indian Institute of Science\\
	Bangalore\\
	\email{jp@ee.iisc.ernet.in}           
	\and
	Jeena R.S.\at
	Dept. of Electronics and Communication Engineering\\
	College of Engineering\\
	Trivandrum\\
	\email{jeena\_rs@yahoo.com
	}}
%
%
\maketitle

\abstract{The revolutionary developments in the field of supervised machine learning have paved way to the development of CAD tools for assisting doctors in diagnosis. Recently, the former has been employed in the prediction of neurological disorders such as Alzheimer's disease. We propose a CAD (Computer Aided Diagnosis tool for differentiating neural lesions caused by CVA (Cerebrovascular Accident) from the lesions caused by other neural disorders by using Non-negative Matrix Factorisation (NMF) and Haralick features for feature extraction and SVM (Support Vector Machine) for pattern recognition. We also introduce a multi-level classification system that has better classification efficiency, sensitivity and specificity when compared to systems using NMF or Haralick features alone as features for classification. Cross-validation was performed using LOOCV (Leave-One-Out Cross Validation) method and our proposed system has a classification accuracy of  over 86\%. }

\section{Introduction}
Cerebrovascular accident (CVA) or cerebrovascular  insult  (CVI), commonly referred to as ``stroke'' is the leading cause of death, next to  ischaemic heart disease and the leading cause of adult disability worldwide \cite{f1,f2}. Globally, 15 million people suffer from stroke every year and of this, a third dies and half of the remaining struggle with permanent disabilities \cite{f5}. The statistics is no different for developing countries like India \cite{f6,f9,f8}. In Trivandrum, the capital of the Indian state of Kerala, the incidence rate of stroke is 135.0 and 138.0 per 1,00,000 inhabitants in the urban community and the rural community respectively per year\cite{f9}. Clearly, stroke has transformed from a disease pertaining to developed nations to a global hazard.

It was these facts which motivated us to develop a prediction system for CVA. As a predecessor to the proposed CAD (Computer Aided Diagnosis) tool for prediction, we have developed a CAD tool for differentiating brain MRIs of subjects who have suffered from stroke (these MRIs will be hereafter referred to as ``stroke MRI'') from the MRI of subjects suffering from other neural disorders (hereafter referred to as ``non-stroke MRI''), which can be diagnosed from structural MRIs. To the best of knowledge of the authors, this is a maiden work in the field of stroke prediction using neuroimaging, though there are several similar works in the literature, pertaining to other neurological disorders, such as schizophrenia \cite{n1}, Alzheimer's disease \cite{40} etc.

In this work, we have used the Haralick features \cite{a6} and Non-negative Matrix Factorisation \cite{17} for feature extraction and SVM for classification. Also, we have introduced a computationally efficient method for combining feature vectors which are linear in two different kernel spaces. For this, we have made use of the distance of the feature vector from the hyperplane as a measure of confidence value of classification, thus improving the classification efficiency obtained by using either one of the two feature vectors individually.

The rest of the paper is organized as follows: Section \ref{s1} gives the details about the database used in this work. Section \ref{s2} briefly discusses the various features used for classification. Section \ref{ss} describes multi-level classification approach. Performance metrics used is discussed in Section \ref{s3}. Finally, the results are given in Section \ref{s4}.
\section{Database}
\label{s1}
The database used for this work is ``The Whole Brain Atlas'', developed by Keith Johnson, MD, and Alex Becker, PhD., with the support of  the Brigham and Women's Hospital Departments of Radiology and Neurology, Harvard Medical School, the Countway Library of Medicine, and the American Academy of Neurology. The database includes the MRI images of neurological disease such as:
\begin{enumerate}
\item Neoplastic Disease (brain tumor)
\begin{itemize}
\item Metastatic adenocarcinoma
\item Metastatic bronchogenic carcinoma
\item Meningioma
\item Sarcoma
\end{itemize}
\item Degenerative Disease
\begin{itemize}
\item Alzheimer's disease
\item Huntington's disease
\item Motor neuron disease
\item Cerebral calcinosis
\end{itemize}
\item Inflammatory or Infectious Disease
\begin{itemize}
\item Multiple sclerosis
\item AIDS dementia
\item Creutzfeld-Jakob disease
\item Cerebral Toxoplasmosis
\end{itemize}
\item Cardiovascular Accident (CVA)
\begin{itemize}
\item Acute stroke: Speech arrest 
\item Acute stroke: ``alexia without agraphia'' 
\item Subacute stroke: ``transcortical aphasia'' 
\item Chronic subdural hematoma
\item Hypertensive encephalopathy, and
\item Cerebral hemorrhage.
\end{itemize}
\end{enumerate}
\begin{figure}
\hfil
\subfigure[Original MRI]{
\includegraphics[width=1in]{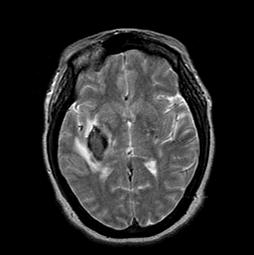}}
\hfil
\subfigure[{\hspace{-0.038cm}Preprocessed MRI}]{
\includegraphics[width=1in]{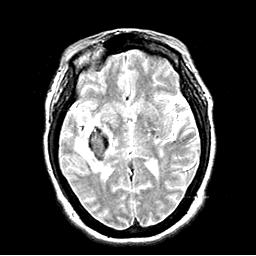}}
\caption{An example  of preprocessing the MRI image}
\label{5}
\end{figure}
A total of 30 T2 weighted MRI images from the database were used in the work.  Out of these, 16 were non-stroke MRIs and 14 were stroke MRIs. All images were manually orientation correccted, cropped and resampled to a resolution of $256 \times 256$. Prior to feature extraction, the images were subjected to normalisation. Normalising the images with respect to the maximum pixel intensity value may introduce noise \cite{40}, so the maximum pixel intensity was considered to be the mean of the highest $0.1 \%$ pixel intensity values, as proposed in \cite{40} and \cite{a7}. The images were further processed for reducing noise using a bilayer filter \cite{lbs1, lbs2}. 
Original and preprocessed images are shown in Fig. \ref{5}.
\section{Feature Extraction}
\label{s2}
We have relied on two sets of features for classification: (1) textural features and (2) features extracted using NMF. 
\subsection{Textural features}
 The textural features introduced by Robert M. Haralick in the paper titled ``Textural features for image classification'' published in the year 1973 \cite{a6} is one of the most important feature extraction techniques for images when the texture of the images has a pivotal role.. These features are generic in nature, meaning they are not developed for a specific imaging modality but can be used for a wide range of images, including biomedical images, where texture is an important property due to the intrinsic spatial tonal variations. The features are computed from matrix referred to as the gray-tone spatial dependence matrix, given by:
\begin{equation}
\mathbf{G}=\left[
\begin{array}{cccc}
c(1,1) & c(1,2) & \cdots & c(1,N_g) \\
c(2,1) & c(2,2) & \cdots & c(2,N_g) \\
\vdots & \vdots & \ddots & \vdots   \\
c(N_g,1) & c(N_g,2) & \cdots & c(N_g,N_g) \\
\end{array}
\right]
\end{equation} 

\begin{figure}
\centering
\subfigure[]{
	\includegraphics[width=.3in]{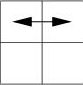}}
\subfigure[]{
	\includegraphics[width=.3in]{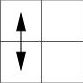}}
\subfigure[]{
	\includegraphics[width=.3in]{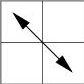}}
\subfigure[]{
	\includegraphics[width=.3in]{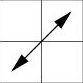}}
 \caption{Four different directional adjacencies possible. (a) Horizontal. (b) Vertical . (c) Left diagonal (d) Right diagonal}
 \label{2}
 \end{figure}

 For an image with $N_g$ gray levels, the matrix will be a square matrix of dimension $N_g$. The element in location $\{i,j\}$ is the count of number of pixels with value $i$ in the neighbourhood of a pixel with value $j$. This can be extended to a probability matrix by dividing with the appropriate count. As shown in Fig. \ref{2}, neighbourhood can be defined in 4 different ways \textit{viz.} horizontal, vertical, left diagonal and right diagonal. Hence, 4 different gray-tone spatial dependence matrices are possible for each image.  
 
  14 different statistics are computed from each of the four gray-tone spatial dependence matrix using the following relations:  
 \begin{enumerate}
 \item Angular Second Moment:
 \begin{equation}
  \sum_i\sum_jc(i,j)^2
 \end{equation}
 \item Contrast: \begin{equation}
 \sum_{n=0}^{N_g-1}n^2\{\sum_{i=1}^{N_g}\sum_{j=1}^{N_g}c(i,j)\},|i-j|=n
 \end{equation}
 \item Correlation: \begin{equation}
 \frac{\sum_i\sum_j(ij)c(i,j)-\mu_x\mu_y}{\sigma_x\sigma_y}
 \end{equation}
 \item Sum of Squares: Variation: \begin{equation}
 \sum_i\sum_j(i-\mu)^2c(i,j)
 \end{equation}
 \item Inverse Difference Moment: \begin{equation}
 \sum_i\sum_j\frac{1}{1+(i-j)^2}c(i,j)
 \end{equation}
 \item Sum Average: \begin{equation}
 \sum_{i=2}^{2N_g}ip_{x+y}(i)
 \end{equation}
 \item Sum Variance: \begin{equation}
 \sum_{i=2}^2N_g(i-f_s)^2p_{x+y}(i)
 \end{equation}
 \item Sum Entropy:\begin{equation}
  -\sum_{i=2}^{N_g} p_{x+y}(i)\log\{p_{x+y}(i)\}
 \end{equation}
 \item Entropy:\begin{equation}
  -\sum_i\sum_jc(i,j)\log (c(i,j))
 \end{equation}
 \item Difference Variance:\begin{equation}
  \sum_{i=0}^{N_g-1}i^2p_{x-y}(i)
 \end{equation}
 \item Difference Entropy: \begin{equation}
 -\sum_{i=0}^{N_g-1}p_{x-y}(i)\log\{p_{x-y}(i)\}
 \end{equation}
 \item Information Measure of Correlation 1: \begin{equation}
 \frac{HXY-HXY1}{\max\{HX,HY\}}
 \end{equation}
 \item Information Measure of Correlation 2: \begin{equation}
 (1-\exp[-2(HXY2-HXY)])^{0.5}
 \end{equation}
 \item Maximal Correlation Coefficient: \begin{equation}
 \text{second largest eigen value of Q}^{0.5}
 \end{equation}
 \end{enumerate}
 \qquad where $\mu_x$,$\mu_y$, $\sigma_x$ and $\sigma_y$ are the mean and standard deviations of $p_x$ and $p_y$, the probability density functions. ${Q}(i,j)$ is given by the following relation:  
 \begin{equation}
 {Q}(i,j)=\sum_k\frac{c(i,j)c(j,k)}{p_x(i)p_y(k)}.
 \end{equation}
 
Since each gray-tone spatial dependence matrix corresponds to a specific directions, the features obtained may vary with the change in the orientation of the image. To have some level of orientation independence, we took the mean and range of the features, hence giving only a total of 28 features per image instead of 56. 
 \subsection{Non-negative Matrix Factorisation}
 
One of the major drawbacks with feature extraction techniques such as PCA, is that negative values of the basis vectors are quite difficult for being interpreted in many practical applications. One of remedies for this is to have a representation where non-negativity is imposed by some means. Given a non-negative matrix $\boldsymbol{A}$, the task is to decompose the matrix into two non-negative matrices $\boldsymbol{V}$ and $\boldsymbol{H}$ subjected to the condition that the Forbenius norm between the given matrix $\boldsymbol{A}$ and the product of the two vectors $\boldsymbol{V}\boldsymbol{H}$ is minimum. 
\begin{equation}
\min_{\boldsymbol{V}\ge0,\boldsymbol{H}\ge0}||\boldsymbol{A}-\boldsymbol{V}\boldsymbol{H}||^2_F
\label{12}
\end{equation}
This decomposition is known as  Non-Negative Matrix Factorization (NMF). The matrix $\boldsymbol{V}$ is called the basis matrix or the mixing matrix and $\boldsymbol{H}$ represents unknown or hidden sources. The beauty of (\ref{12}) lies in the perspective of viewing each column of $\boldsymbol{A}$ as a linear combination of columns of $\boldsymbol{V}$ with the weights given by the components of each column in $\boldsymbol{H}$ \cite{nn1,nn2,nn3}. The number of columns in $\boldsymbol{V}$ is very much less than the number of rows in $\boldsymbol{A}$. Though NMF seems to be a computationally complex operation, several algorithms have been developed in the last decade for the computationally efficient implementation of NMF.

The maiden work in the field of NMF (Non-negative Matrix Factorisation) can be traced back to a 1994 paper by Paatero and Tapper \cite{17} in which they performed factor analysis
on environmental data \cite{f35}. Their aim was to find the common latent features or latent variables that explained the given set of observation vectors. Some elementary variables combine together positively to give each of the factor.  A factor can either be present, in which case it has a positive effect or the factor can be absent, in which case the factor has null influence. Clearly, there is no room for a ``negative'' influence and hence  it often makes sense to
constrain the factors to be non-negative.

 \section{ Multilevel SVM Classification}
 \label{ss}
 SVM is a widely used supervised machine learning algorithm for classification developed by Vladimir N. Vapnik and the current
 standard incarnation (soft margin) was proposed by Vapnik and Corinna Cortes in 1995 \cite{34}. An in-depth discussion on SVM classification can be found in \cite{n2}. In our proposed multi-level classification, two support vector models are created using NMF features and Haralick features. Now, given a feature vector $\Phi(\overrightarrow x)$, in the feature space $\Phi(.)$, we compute a score based on its distance from the decision boundary hyperplane $f(x)$ as,\cite{n3}
 \begin{equation}
 \text dist (\Phi(x),f(x))=\frac{f(x)}{\sum\limits_{i \in SV}|y_i\alpha_i\Phi(x)|^2}
 \end{equation} 
 where $y_i\in \{-1,1\}$, $\alpha_i : $ vector weights for support vectors.
 
 The scores provide an estimate of how good the classification is.  Larger the score, larger will be the distance from the hyperplane and hence higher will be probability of the sample to lie in that class \cite{n3}. In our work, for a given test sample, we compute the scores for both the support vector models. The model which gives the highest absolute value for the score is assumed to have classified the sample correctly. This approach improved the accuracy of the classification.
 \section{Performance Metrics}
 \label{s3}
 Three performance metrics were used in this work:
 \begin{enumerate}
 \item Sensitivity ($SN$), which is the measure of the system's ability to identify stroke MRIs.
 \item Specificity ($SP$), which is a measure of the system's ability to identify non-stroke MRIs.
 \item Accuracy ($AC$), which is a measure of the system's net classification efficiency.
 \end{enumerate}
 Before defining these metrics mathematically, we introduce the following terms:
 \begin{itemize}
 \item $TP$: True Positive, stroke MRI identified as stroke MRI.
 \item $TN$: True Negative, non-stroke MRI identified as non-stroke MRI.
 \item $FP$: False Positive, non-stroke MRI identified as stroke MRI.
 \item $FN$: False Negative, stoke MRI identified as non-stroke MRI.
 \end{itemize}
 Now, we can mathematically define sensitivity, specificity and accuracy as:
 \begin{equation}
 SN=\frac{TP}{TP+FN}
 \end{equation}
 \begin{equation}
 SP=\frac{TN}{TN+FP}
 \end{equation}
 \begin{equation}
 AC=\frac{TP+TN}{TN+TP+FP+FN}
 \end{equation}
    \begin{figure}[h!]
     \centering
     \includegraphics[width=3in]{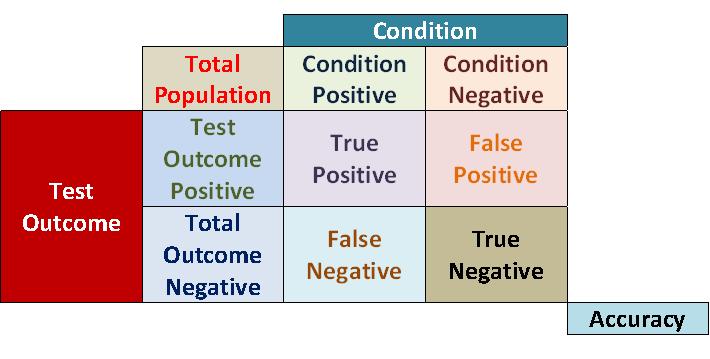}
     \caption{Sample confusion matrix}
     \label{cf1}
     \end{figure}
 The confusion matrix used in this work is shown in Fig. \ref{cf1}.
 
 \section{Results}
 \label{s4}
 LOOCV (Leave-One-Out Cross Validation), which is a special case of LpOVC (Leave-p-Out Cross Validation) was used for cross validation. The steps adopted were,
 \begin{enumerate}
 \item \textbf{Step 1: }Out of the $N$ samples with ground truth, $N-1$ were chosen as the training set and remaining one was chosen as the test data. Using this set of training data, both classification models were generated and using these models, the test data was classified based on the SVM score. Depending on the classification output and the ground truth, one of $TP$, $TN$, $FP$ or $FN$ was incremented.
 \item \textbf{Step 2: }In the next iteration, a different sample was chosen as the test data and Step 1 is repeated. 
 \item \textbf{Step 3: }This process is continued until all the samples have been used as a test data once.
 \end{enumerate}

    \begin{figure}
    \centering
    \subfigure[Confusion matrix when  NMF weight vector alone is used as the features]{
    \includegraphics[width=3in]{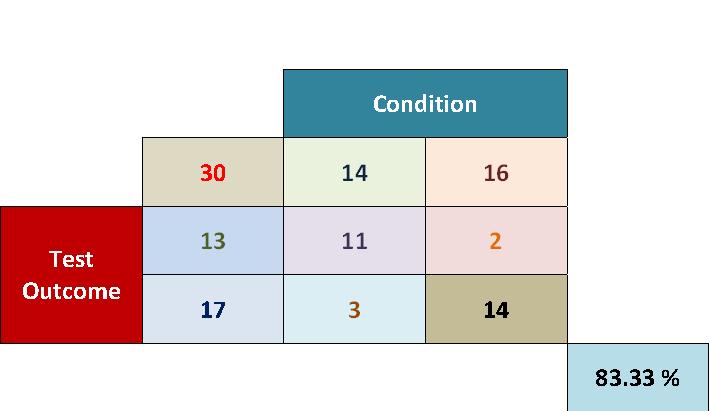}}
    \centering
    \subfigure[Confusion matrix when  Haralick features alone is used]{
    \includegraphics[width=3in]{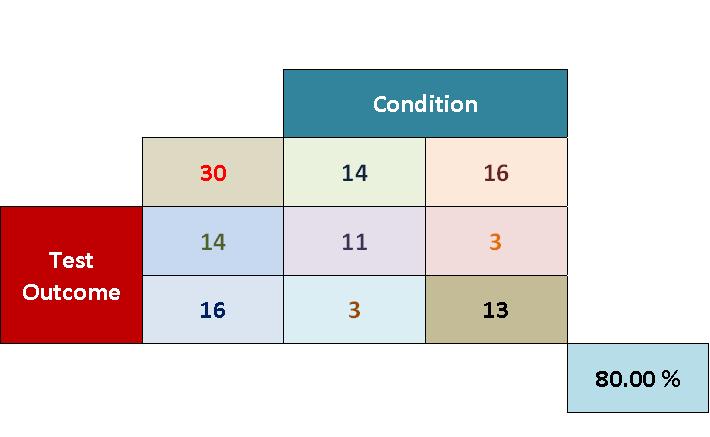}}
    \caption{Confusion matrices for various features considered individually.}
    \label{c5}
    \end{figure}
        \begin{figure}
        \centering
        \subfigure[Confusion matrix obtained when concatenated Haralick feaures and NMF weight vector is used as the feature vector]{
        \includegraphics[width=3in]{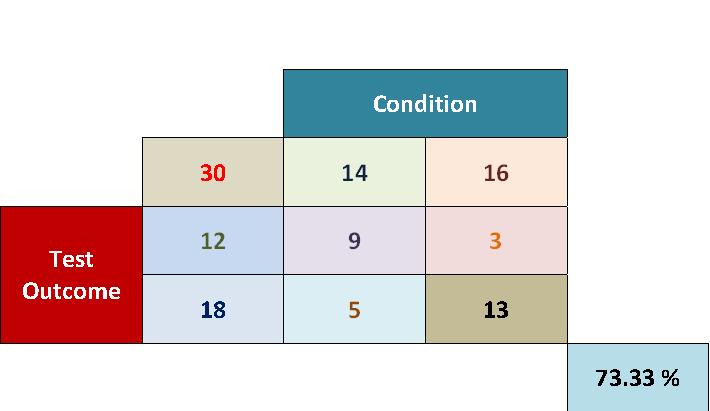}
        \label{w1}}
        \centering
        \subfigure[Confusion matrix when multi-level SVM is used]{
        \includegraphics[width=3in]{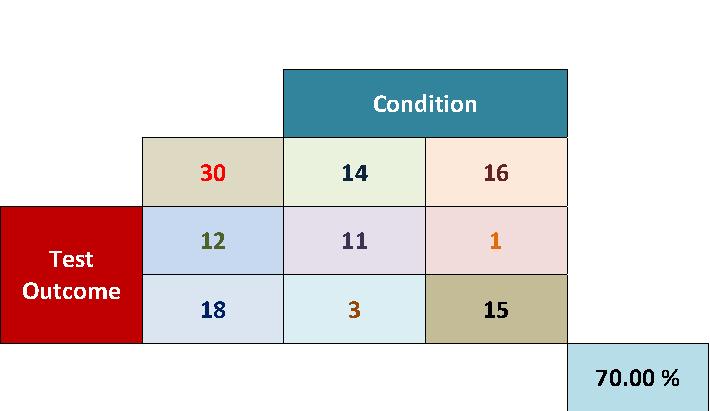}
        \label{w2}}
        \caption{Confusion matrices for various features considered simultaneously.}
        \label{c6}
        \end{figure}
          \begin{figure}[h!]
                \centering
                \includegraphics[width=3in]{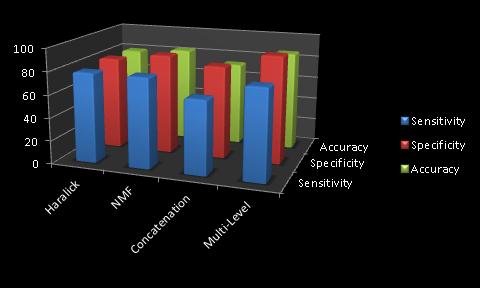}
                \caption{Comparison of multi-level SVM with simple SVM}
                \label{cfe}
                \end{figure}
         
                      \begin{table}
                      	\caption{Classification results when  Haralick features alone is used}
                      	\label{q1}       
                      	%
                      	%
                      	\begin{tabular}{p{2cm}p{2cm}p{2cm}p{0.8cm}}
                      		\hline\noalign{\smallskip}
                      		& Linear & MLP & RBF \\
                      		\noalign{\smallskip}\svhline\noalign{\smallskip}
                      		 Sensitivity & 71.43 & 78.57 & 71.00 \\ 
                      		Specificity & 87.50 & 81.25 & 81.25 \\ 
                      		Accuracy & 80.00 & 80.00 & 76.67 \\
                      		\noalign{\smallskip}\hline\noalign{\smallskip}
                      	\end{tabular}
                      \end{table}
                      
                                \begin{table}
                                	  \caption{Classification results when  NMF weight vector alone is used as the features}
                                  \label{q2}       
                                	%
                                	%
                                	\begin{tabular}{p{2cm}p{2cm}p{2cm}p{0.8cm}}
                                		\hline\noalign{\smallskip}
                                		& Linear & MLP & RBF \\
                                		\noalign{\smallskip}\svhline\noalign{\smallskip}
                                		Sensitivity & 78.57 & 64.29 & 71.43 \\ 
                                	Specificity & 87.50 & 87.50& 81.25 \\ 
                                	Accuracy & 83.33 & 76.67 & 76.67 \\
                                		\noalign{\smallskip}\hline\noalign{\smallskip}
                                	\end{tabular}
                                \end{table}

                                \begin{table}
                                                              	 \caption{Comparison of multi-level SVM with simple SVM}
                                                              \label{q3}      
                                                              	%
                                                              	%
                                                              	\begin{tabular}{p{2cm}p{2cm}p{2cm}p{2cm}p{1.5cm}}
                                                              		\hline\noalign{\smallskip}
                                                              	& Haralick & NMF & Concatenated & Multi-Level  \\
                                                              		\noalign{\smallskip}\svhline\noalign{\smallskip}
                                                              	Sensitivity & 78.57 & 78.57 & 64.29 & \textbf{78.57}  \\ 
                                                              	Specificity & 81.25 & 87.50& 81.25 & \textbf{93.75}\\ 
                                                              		Accuracy & 80.00 & 83.33 & 73.33 & \textbf{86.67}\\
                                                              		\noalign{\smallskip}\hline\noalign{\smallskip}
                                                              	\end{tabular}
                                                              \end{table}

    TABLE 1 and II shows the result of using NMF and Haralick features independently for classification using different kernel functions. For MLP, the parameter was $[1 -9]$ and for RBF, the sigma value was $40$. These values were found to give the best results. 
    
    By using the proposed multi-level classification, we could improve the accuracy, sensitivity and specificity rather than by just concatenating the features. The results are shown in TABLE III, Fig. 5 and Fig. 6.
 \section{Conclusion}
 In this paper, we have proposed a novel CAD tool that can classify the structural MRI images of patients who had stroke from the images of patients suffering from other neurological disorders. By incorporating a multi-level classification system, we have achieved a classification efficiency of more than 83\% by using Non-negative matrix factorisation and Haralick features as the features and SVM as the classifier. This can be further improved by incorporating other features. We believe that this can be extended to a CAD systems for predicting CVAs.
\section*{Acknowledgement}
The authors place on record their deepest gratitude to
 Dr. Shine M. Babu, Assistant Professor, Dr. Somervell Memorial CSI Medical College \& Hospital, Trivandrum and Dr. Benedict Bright, Medical Officer, CSI Mission Hospital, Trivandrum for the support they extended to this work.
\bibliographystyle{splncs} 
\bibliography{thes}
\end{document}